\documentclass{article}

\usepackage{PRIMEarxiv}

\usepackage[utf8]{inputenc} 
\usepackage[T1]{fontenc}    
\usepackage{hyperref}       
\usepackage{url}            
\usepackage{booktabs}       
\usepackage{amsfonts}       
\usepackage{nicefrac}       
\usepackage{microtype}      
\usepackage{lipsum}
\usepackage{fancyhdr}       
\usepackage{graphicx}       
\graphicspath{{media/}}     

\title{A Novel Home-Built Metrology to Analyze Oral Fluid Droplets 
and Quantify the Efficacy of Masks
}

\author{
  Ava Bhowmik \\
  The Harker School \\
  San Jose, CA\\
  \texttt{26avab@students.harker.org} \\
}

\begin{document}
\maketitle
\begin{abstract}
Wearing masks is crucial to preventing the spread of potentially pathogen-containing droplets, especially amidst the COVID-19 pandemic. However, not all face coverings are equally effective and most experiments evaluating mask efficacy are very expensive and complex to operate. In this work, a novel, home-built, low-cost, and accurate metrology to visualize orally-generated fluid droplets has been developed. The project includes setup optimization, data collection, data analysis, and applications. The final materials chosen were quinine-containing tonic water, 397-402 nm wavelength UV tube lights, an iPhone and tripod, string, and a spray bottle. The experiment took place in a dark closet with a dark background. During data collection, the test subject first wets their mouth with an ingestible fluorescent liquid (tonic water) and speaks, sneezes, or coughs under UV darklight. The fluorescence from the tonic water droplets generated can be visualized, recorded by an iPhone 8+ camera in slo-mo (240 fps), and analyzed. The software VLC is used for frame separation and Fiji/ImageJ is used for image processing and analysis. The dependencies of oral fluid droplet generation and propagation on different phonics, the loudness of speech, and the type of expiratory event were studied in detail and established using the metrology developed. The efficacy of different types of masks was evaluated and correlated with fabric microstructures. All masks blocked droplets to varying extent. Masks with smaller-sized pores and thicker material were found to block the most droplets. This low-cost technique can be easily constructed at home using materials that total to a cost of less than 50 USD. Despite the minimal cost, the method is  accurate and the data is quantifiable.

\end{abstract}

\section{Introduction}
The COVID-19 pandemic is the most disastrous global health crisis since the 1918 influenza pandemic [Feehan, 2021]. As of December 30, 2021, the SARS-CoV-2 virus has infected over 285 million people and caused almost 5.42 million fatalities worldwide [Johns Hopkins, 2021]. In addition to COVID-19, which has been the subject of much discussion recently, the influenza virus infects around one billion people worldwide and inflicts close to 650,000 deaths annually [WHO, 2019]. These and many other contagious respiratory diseases spread through exposure to the corresponding pathogens in small saliva or mucus droplets generated by infected individuals when they breathe, talk, cough, or sneeze. Inhalation of such particles often causes infection. The propagation of these infectious droplets can be obstructed by the wearing of an effective mask. 
While standard masks such as N95 respirators are proven to be extremely efficient at filtering airborne particles, top-grade personal protective equipment (PPE) should be reserved for medical first responders and other healthcare workers, especially amidst PPE shortages. On top of that, surgical masks are not reusable and are made out of non-recyclable plastics that are harmful to the environment [Alexander, 2021]. Thus, the general public is encouraged by the Centers for Disease Control and Prevention (CDC) to wear alternative face coverings such as homemade or commercially available fabric masks [CDC, 2021]. However, not every mask can block droplets to the same degree, and currently there is no easy and low-cost method for determining the efficacy of specific masks. Accurate experimental setups that have been reported require a lab environment, complex equipment, and experience to operate. It is both costly and impractical to test every single kind of mask in a lab setting. In addition, these techniques are generally not accessible and repeatable in home environments.
In this work, a novel, home-built, low-cost, and accurate metrology to visualize orally generated fluid droplets and consequently determine the efficacy of masks has been developed. This method is the world’s first fluorescence-based technique for oral droplet visualization that utilizes common household materials. It is based on the fluorescent properties of a common soft drink, tonic water, to make droplets generated by expiratory events visible under UV darklight. The fluorescent emission from the oral fluid droplets for different expiratory events are captured in slow-motion using a smartphone camera and processed and analyzed using open-source softwares. The results of these analyses are applied to determine the efficacy of masks made from various materials.

\section{Prior Work}
\label{sec:headings}

Many of the past experiments that have been reported for visualizing oral fluid droplets are accurate but generally expensive and complex to operate. The table shown in Figure 1 presents a summary of four representative past works concerning the detection and/or visualization of respiratory droplets and aerosols.

A previously reported test setup that is potentially feasible at home uses colored dyes to quantify the amount of droplets generated by an expiratory event [Mythbusters, 2011]. However, this system has several limitations. The microdroplets cannot be recorded while airborne due to the low signal to noise ratio, thus eliminating the possibility of studying the trajectory and propagation of the particles. The dye droplets would also become distorted when landing on the ground, causing some stains to merge and others to distort. Additionally, the dyes used could potentially be toxic to ingest or unpleasant to put in the mouth.

One researcher utilized a light scattering technique combined with high speed videography in order to visualize the cloud of saliva and mucous droplets expelled by a sneeze [Bourouiba, 2016]. This method is fairly accurate and can capture fast-moving droplets, but any foreign particles present in the test chamber would also scatter light. In order to address this problem, a high-efficiency particulate air (HEPA) filter is used, although the filtration system is not fully effective and dust could still enter the test chamber. Furthermore, the wavelength of light scattered by the droplets is the same as the wavelength of the source light, thus lowering the signal to noise ratio for detection. 

Another experiment utilized a laser light sheet rather than a large-area light illumination to eliminate the issue of the source light also being captured by the camera [Anfinrud, 2020]. However, this technique can only visualize droplets passing through the thin light sheet at any given moment. In addition, the expensive equipment such as HEPA filters, high-intensity lasers and lights, and high-speed cameras render these experiments inoperable and inaccessible in home environments. Also, if not handled properly, the powerful laser can cause eye damage and visual impairment.

Simpler methods for determining mask efficacy, such as the “candle test” or testing for a hydrophobic coating, are not quantifiable or accurate. In the example of the candle test, in which a subject attempts to blow out a candle while wearing a mask, outside variables, such as the type of candle and personal lung strength can affect the outcome [Baskar, 2020]. In addition, blowing may not be a reliable proxy for small aerosols exiting with normal speaking or coughing.

\begin{figure}[h]
\vspace{1mm}
\includegraphics [scale=0.35]{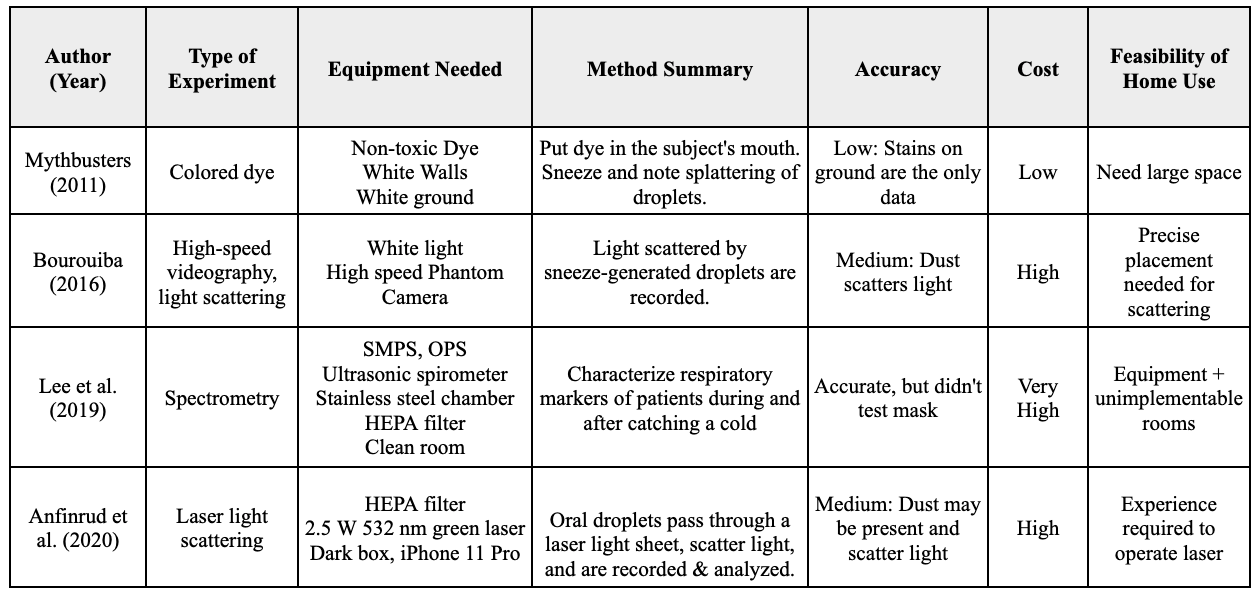}
\centering
\caption{\small Summary of four selected representative past works for oral droplet visualization and analysis. The only experiment setup that is possibly viable in a home environment is not very accurate and quite messy [Mythbusters, 2011]. The rest of the experiments require complex and expensive equipment like spectrometers, HEPA filters, high speed cameras, and high-power lasers, and thus are not suitable for home use.}
\end{figure}

\section{Methods and Materials}
\label{sec:headings}

In this research, a novel,  accurate, repeatable, home-built, and low-cost metrology to analyze oral fluid droplets and evaluate the efficacy of specific masks has been developed based on fluorescence, high-speed videography, and image processing and analysis techniques. This experiment consisted of four major steps: setup optimization, data collection, data processing and analysis, and applications, as illustrated in Figure 2.

\begin{figure}[h]
\centering
\includegraphics [scale=0.42]{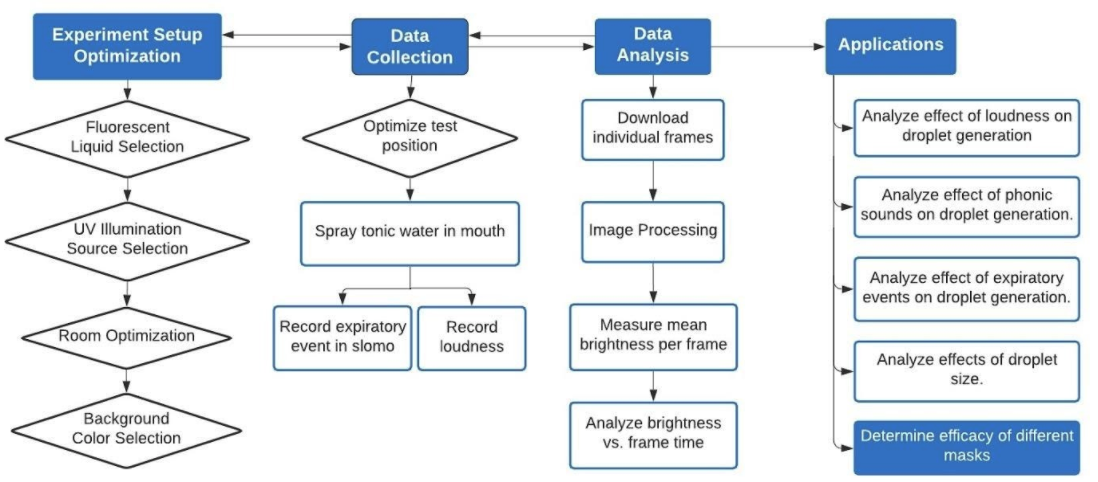}
\caption{\small The four stages of this metrology: setup optimization, data collection, data analysis, and applications. For setup optimization, the fluorescent liquid, UV light source, and setting conditions were tested and perfected. During data collection, recordings of the fluorescent droplets are taken with a smartphone. The data is then processed and analyzed. This metrology can be applied to a broad range of applications, including the study of droplet propagation, mask testing, and education purposes.}
\vspace{2 mm}
\end{figure}

\begin{figure}[h]
\vspace{1mm}
\includegraphics [scale=0.428]{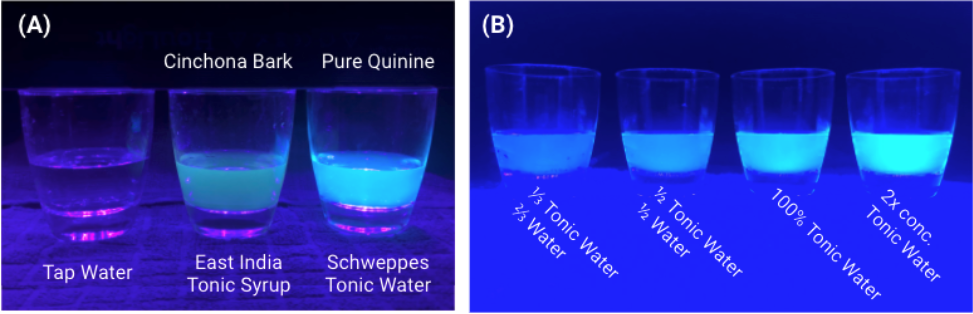}
\centering
\caption{\small Comparison of (A) different quinine containing liquids and tap water (control) and (B) different concentrations of tonic water. Between tonic syrup, which contains 5x the quinine concentration of tonic water, and Schweppes tonic water, the tonic water appeared the brightest. This is because tonic syrup contains cinchona bark, which quinine is extracted from, but not pure quinine.
}
\end{figure}

\paragraph{Setup Optimization}  In this step of the process, the independent variables such as fluorescent liquid choice, UV light source selection and configuration, and setting conditions were optimized. Repeated testings and observations were performed to select the optimal materials and arrangements.

An ingestible, fluorescent liquid that has a similar viscosity to saliva and fluoresces brightly enough to be seen and captured using an iPhone camera needed to be selected for the experiment. After studying a wide range of household products, tonic water was the only one that fulfilled all the criteria. Tonic water contains a fluorophore called quinine. The excitation wavelength range of quinine is approximately 270 nm to 400 nm and the emission spectrum is approximately 380 nm to 530 nm, giving tonic water its signature blue glow under UV dark light (350-400 nm wavelength) [Lawson-Wood, 2018]. In order to further optimize the fluorescent liquid, different quinine-containing drinks and concentrations were evaluated. Figure 3 shows images of the tested liquids along with tap water as a control material. The most notable candidate tested was East India Tonic Syrup, which was advertised to contain 5 times the quinine concentration of regular tonic water. However, as is obvious in Figure 3A, the tonic water was significantly brighter than the tonic syrup. This is because the tonic syrup contains cinchona bark, which pure quinine is extracted from, rather than the substance itself. Thus, Schweppes branded tonic water was determined to be the best quinine-containing liquid. After testing different concentrations of tonic water, regular tonic water was found to be the most suitable concentration for the experiment. It was sufficiently bright, as seen in Figure 3B, and also reduced variability due to quinine decomposition from high temperature.

Between a UV blacklight flashlight (385-395 nm wavelength) and a set of UV darklight party tube lights (397-402 nm wavelength), the optimal light source determined was the tube light. It provides uniform lighting over a large area, whereas the flashlight only lights up a small area very intensely. This would mean that only droplets passing through the center of the beam would be visible. In addition, the party tube light had a wavelength that was rated safer for exposure to human eyes and skin.

Figure 4 shows the setup, which was established in a closet of the home, with the room light off, and a dark background achieved with a black poster board or a towel fixed to the wall. As seen in this figure, the setup consisted of UV tube lights suspended from the closet hanger rod using a string, an iPhone camera on top of a tripod, a spray bottle for calibration purposes, quinine-containing tonic water, and a dark poster board or towel to serve as the black background. Figure 5 shows the alternate setups that were experimented with. The other conditions tested had a white background and/or the room light on. Turning the room light off and having a dark background both increase the signal-to-noise ratio, making the fluorescent droplets more clear and thus data analysis more accurate.

\begin{figure}[h]
\includegraphics [scale=0.431]{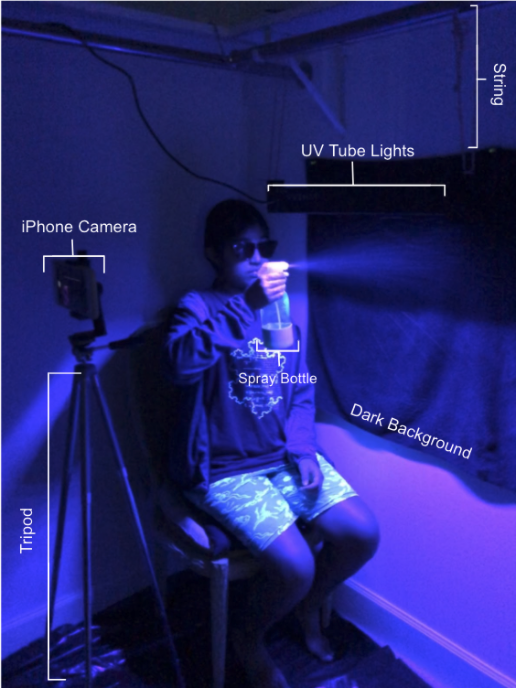}
\centering
\caption{\small The final experimental setup conditions and materials used (tonic water, UV dark light tube lights, string, an iPhone and tripod, a dark background, and a spray bottle). The cost of the setup should total to no more than 50 USD, assuming the person conducting the experiment already had a phone beforehand.
}
\end{figure}

\begin{figure} [h]
\includegraphics [scale=0.431]{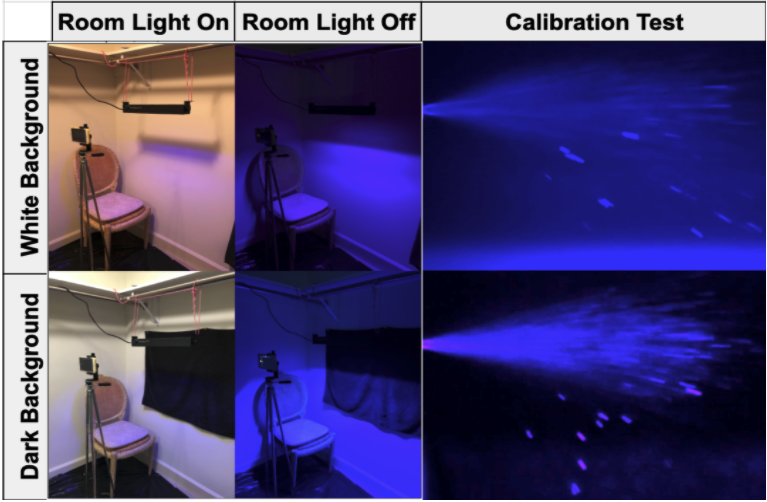}
\centering
\caption{\small Some of the alternate setup conditions tested. The independent variables changed were the background color and whether the room light was on or off. The optimal setting was achieved with the room light off with a dark background. Since the dark background absorbed all of the light that came in contact with it, it drastically increased the signal to noise ratio. This made the droplets clearer and thus image processing more effective and data analysis more accurate.}
\vspace{3mm}
\end{figure}

\paragraph{Data Collection} During data collection, the test subject first lightly wets their mouth with tonic water, preferably using a spray bottle to reduce variability of the amount of tonic water. The iPhone camera should be recording in the slow-motion (“slo-mo”) setting, which is 240 frames per second. Then, the person performs an expiratory event (speaking, sneezing, or coughing). For every trial, the loudness level of speech should remain constant, which was recorded using a loudness measurement application on the iPhone. With this process, the droplets generated were visualized, recorded, and analyzed as described below.

\paragraph{Data Analysis} The video of the spray was separated into individual frames using the open-source and free VLC software package from a non-profit organization [VideoLAN, 2021]. This was performed using VLC’s scene video filter, which can be enabled in the “Preferences” menu. The recording ratio needs to be set to 1 to ensure that every single frame is saved. A folder to store the images will have to be created in advance. Next, the frames were imported into ImageJ, a freeware from the National Institutes of Health (NIH), as an image stack for processing and analysis [NIH, 2021]. In ImageJ, the frames are enhanced by adjusting the histogram. The final products of the image enhancement can be seen in Figure 6. The image montage shows the progression of the microdroplet cloud generated by the word “Fruits” being spoken loudly at about 92 decibels (dB). Then, using the “Measure…” function on the ImageJ software menu, the mean brightness values of each frame are collected. They can subsequently be graphed in an intensity vs. time plot to quantify the generation and dissipation of droplets and aerosols, as narrated in the Results and Discussion section below.

\begin{figure}[h]
\includegraphics [scale=0.431]{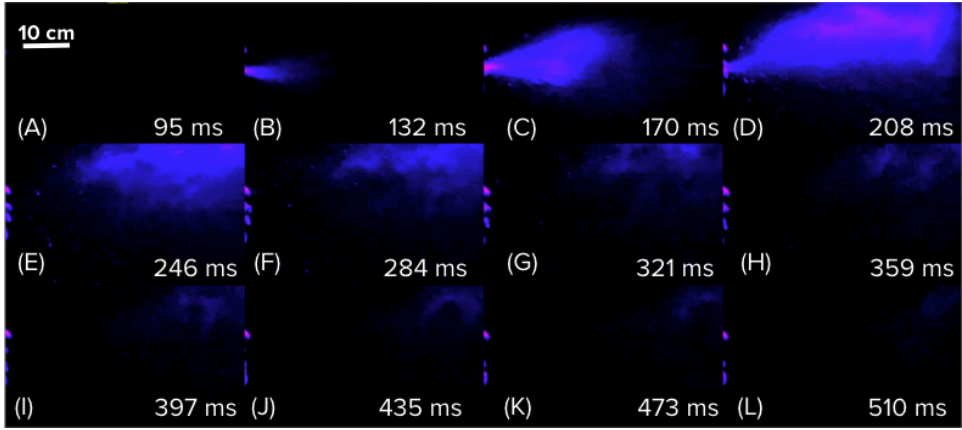}
\centering
\caption{\small Propagation of microdroplets generated by the word “Fruits” spoken loudly, shown in increments of 38 milliseconds. This technique is able to capture the microdroplets lingering in the air long after their dispersal. This shows that 6 feet of social distancing alone is not enough to combat COVID-19, as the contagious particles can stay in the air long after a sick individual has left the area. Thus, it is crucial to wear face coverings in order to prevent both dispersal and inhalation of saliva droplets.}
\end{figure}

\section{Results and Discussion}
\label{sec:others}
Using the experimental setup and the procedures described above, the effects of different variables such as the phonic sound, loudness levels, types of expiratory event, and mask-type worn on the amount and propagation characteristics of droplets generated were studied. In this section, the results of these analyses are presented.

\subsection{Droplet Generation and Propagation} The dependency of the droplet generation and propagation on different variables was studied. By testing different spoken words, it was determined that the “F” and “Ph” phonics generated substantially more droplets compared to vowels and other consonants, as can be seen in Figure 7. This is because the vibration of the front teeth against the bottom lip while uttering these sounds propel the droplets outward. The same logic can be applied to the “Th” sound. It was also observed that the amount of fluid expelled during speech increased with the loudness levels of the sound generated. Figure 8A shows the superimposed mean brightness vs. time charts for different loudness levels of speech, and the relationship between microdroplet generation and the loudness levels. By looking at the data, it is clear that louder speech generates more microdroplets. The correlation between loudness level and the peak intensity of the recording is shown in Figure 8B. The higher the loudness level is, the more droplets are generated. This experiment reaffirmed the already well established fact [Jayaweera, 2020] that aerosols, which are droplets smaller than 10 microns in diameter, float in the air for prolonged periods of time and eventually dehydrate, making them particularly infectious (Figure 9). As depicted in this figure, the trajectory of the bigger droplet is steeper and thus travels a shorter distance compared to the relatively smaller droplet. The dehydrated nuclei of the aerosols can stay in the air for minutes or even hours. As can be observed in Figure 10, the droplet cloud expelled by a sneeze lingered for much longer compared to the droplets from other expiratory events (cough and speech). This may be due to the variations in the aerosol to droplet ratio generated by different expiratory events. Since aerosols remain in the air for prolonged periods of time, a higher aerosol to droplet ratio would result in a much longer dissipation time.

\begin{figure}[h]
\includegraphics [scale=0.377]{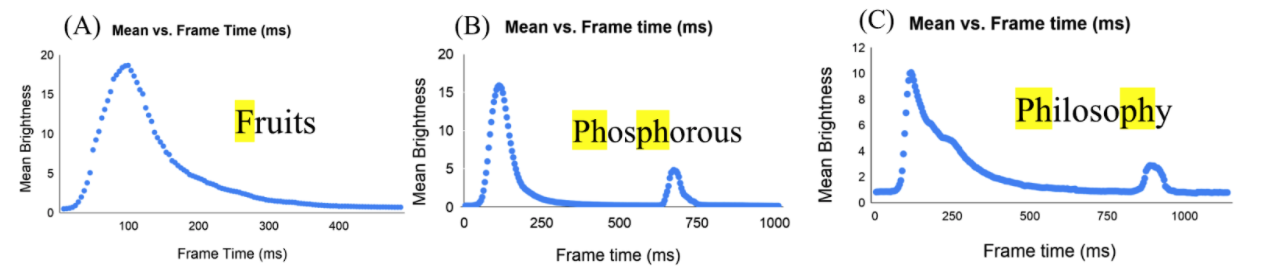}
\centering
\caption{\small The “F” and “Ph” sounds generate more droplets than other phonics, as shown in the peak brightness in all three words above. This is because many saliva droplets are propelled outwards when the front teeth vibrate against the bottom lip while uttering these sounds. As a result, a noticeable cloud of fluid is expelled, making these phonics more dangerous to utter in public because they can easily spread a respiratory pathogen.}
\end{figure}

\begin{figure}[h]
\includegraphics [scale=0.431]{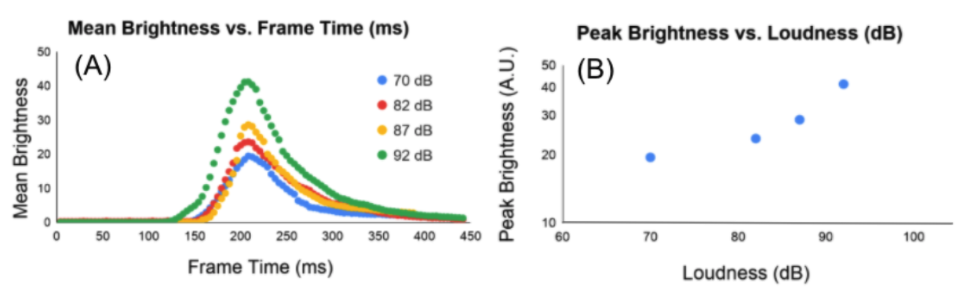}
\centering
\caption{\small Louder speaking generates more droplets, as indicated by brightness levels for different loudness levels of speech. The peak brightness of each recording increases as the speech gets louder. In addition, the droplets generated by louder speech linger in the air for longer than softer speech. This means that not only does louder speech result in greater microdroplet generation, but it also propels droplets further.}
\end{figure}

\begin{figure}[h]
\includegraphics [scale=0.431]{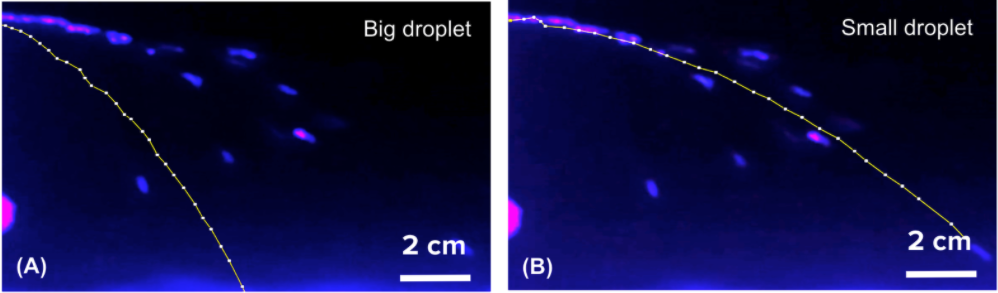}
\centering
\caption{\small This experiment reaffirms the already well-established fact that aerosols, which are droplets less than 10 microns in diameter, linger in the air for prolonged periods of time and remain airborne for much longer than larger droplets. As shown in the figure above, the trajectory of the small droplet goes much further than the large one.}
\end{figure}

\begin{figure}[h]
\includegraphics [scale=0.4]{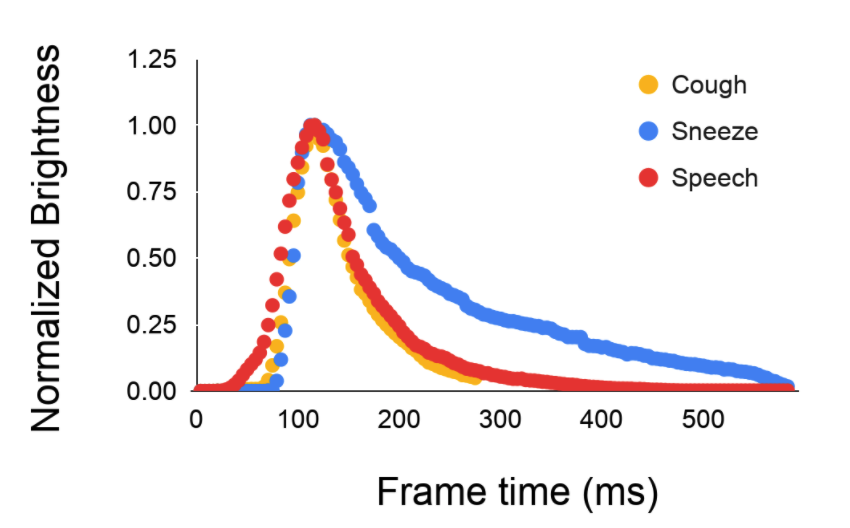}
\centering
\caption{\small The amount of time it takes the droplets to dissipate varies across different oral activities. Sneezing has a significantly longer dehydration time compared to speaking and coughing, making it especially infectious.}
\end{figure}

\subsection{Mask Efficacy Comparison} Finally, the experiments described above were repeated with the subject wearing a different mask each time. The masks tested were made of thin cotton, thick cotton, linen, thin polyester, thick polyester, surgical, and N95. The peak brightness values from the recordings of each trial for a given mask were recorded using the procedures described above. The efficacy of each tested mask was determined and correlated to its nanostructure using averaged peak intensity values and the scanning electron microscopy (SEM) images of different fabrics, which can be seen in Figure 11. The connection can be made that both thicker material and smaller and less numerous pores result in fewer droplets escaping the face covering. As seen in Figure 12, out of all the face coverings tested (both homemade and standard), the N95 mask predictably performed the best. A thick polyester mask was the most effective cloth mask at blocking the droplets. It can be observed that the droplets escaping from the cotton masks are significantly smaller than with the linen face covering, almost a fine mist compared to the distinctive, individual droplets. This may be because when the initial droplets pass through the holes in the cotton fabric, they get broken up into smaller droplets. However, in linen, it is possible some droplets passing through the fabric will merge on the other side. As a result, while cotton masks can help hold back oral droplets, it is possible that wearing them may be even more dangerous than not wearing a mask at all due to droplets that would otherwise fall to the ground quickly being converted into aerosols that linger in the air.

\begin{figure}[!]
\includegraphics [scale=0.41]{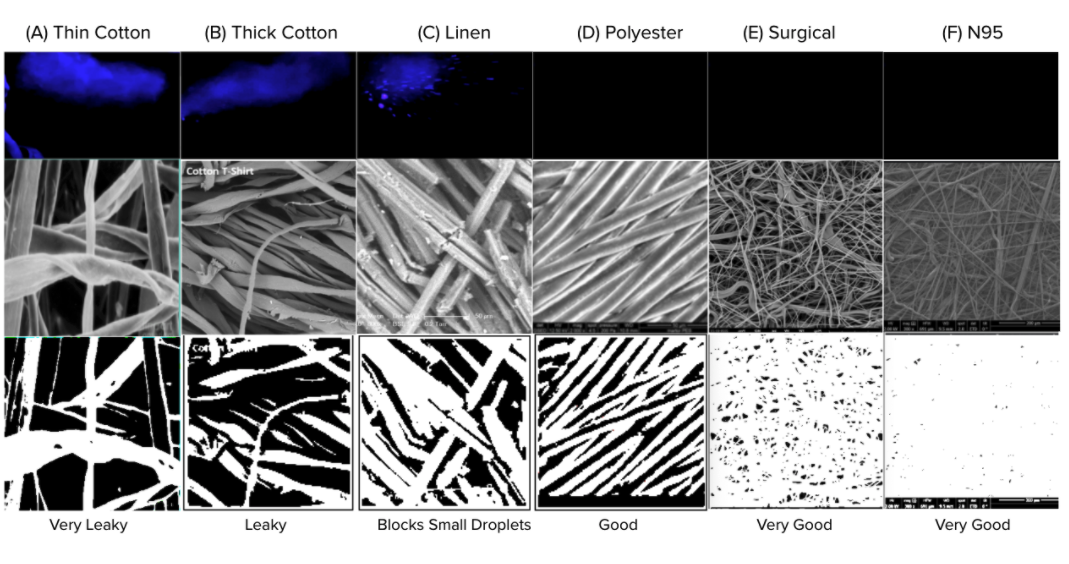}
\centering
\caption{\small The bigger and more numerous pores there are, the more droplets escape. Thinner material is also correlated with more droplets being generated. This figure shows the brightest frame from the videos of each mask  side by side with the scanning electron microscope (SEM) images of the fabric and the thresholded version of the images. Thick and thin cotton, which are both extremely porous, were the least effective. Not only did they allow small droplets to pass through, but they also allowed larger droplets to be broken up into aerosols when they encountered the holes. Linen blocked many aerosols in addition to small and big droplets but still let a considerable amount through. Polyester barely let any droplet through and the surgical and N95 masks were very effective.}
\end{figure}

\begin{figure}[!]
\includegraphics [scale=0.421]{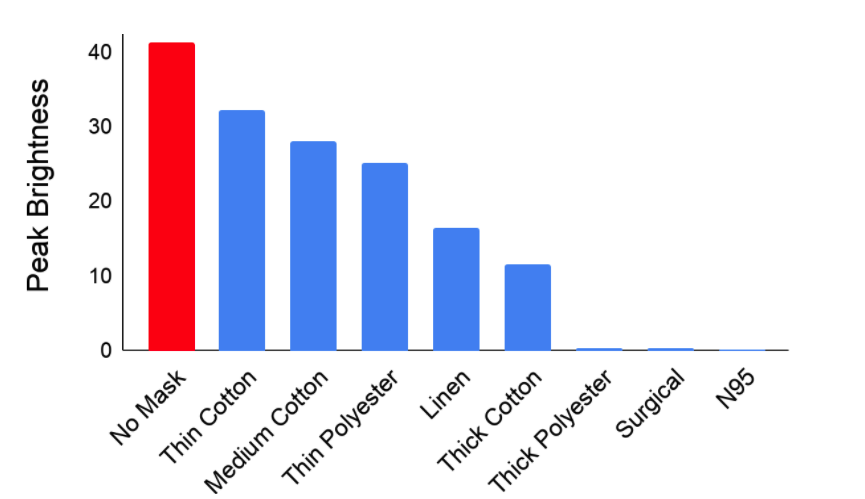}
\centering
\caption{\small After comparing the results for no mask, thin cotton, medium cotton, thick cotton, linen, thin polyester, thick polyester, surgical, and N95 masks, a thick polyester face covering was determined to be the most effective cloth mask, although the N95 mask performed best overall. Wearing any mask was better than none.}
\end{figure}

\subsection{Applications} This metrology can be used to study the generation, propagation, trajectory, and dissipation of orally-generated microdroplets. When data collection is carried out while the test subject is wearing a mask, the efficacy of the face covering can be tested, as demonstrated in this work.

\section{Conclusions}
\label{sec:headings}

In summary, a novel, home-built, low-cost, and accurate metrology to visualize oral fluid droplets and quantify mask efficacy has been developed. This fluorescence-based technique consisted of four major blocks: setup optimization, data collection, data analysis, and applications.

The proposed system can be easily built and used in a home environment and can be carried out in many different settings. The setup is easily scalable once implemented into a portable prototype. Once easily transportable, this technique can be used for educational and mask evaluation purposes around the globe. The use of this method can be further simplified in the future by automating the data processing and analysis processes to make it into a mobile application.

\newpage
\section*{Acknowledgements}
\label{sec:headings}

I would like to thank Dr. Archelle Georgiou for providing valuable inputs and guidance. Additionally, I am grateful for the continuous support and constructive feedback that my parents and brother provided throughout this project.

\end{document}